\documentclass[10pt, a4paper, final, journal, twocolumn, twoside, compsoc]{IEEEtran}

\usepackage{cite}
\usepackage{tabularx,booktabs}

\ifCLASSINFOpdf
  \usepackage[pdftex]{graphicx}
\else
  \usepackage[dvips]{graphicx}
\fi

\usepackage{amsmath}

\usepackage{algorithmic}
\usepackage{comment}
\usepackage{hyperref}

\usepackage{array}
\usepackage[dvipsnames]{xcolor}
\ifCLASSOPTIONcompsoc
 \usepackage[caption=false,font=scriptsize,labelfont=sf,textfont=sf]{subfig}
\else
 \usepackage[caption=false,font=footnotesize]{subfig}
\fi

\newcommand{\Hquad}{\hspace{0.5em}} 
\newcommand{\HHquad}{\hspace{0.05em}}

\usepackage[pscoord]{eso-pic}
\newcommand{\placetextbox}[3]{
  \setbox0=\hbox{#3}
  \AddToShipoutPictureFG*{
    \put(\LenToUnit{#1\paperwidth},\LenToUnit{#2\paperheight}){\vtop{{\null}\makebox[0pt][c]{#3}}}%
  }%
}%

\begin{document}
\placetextbox{0.5}{1}{This is the author's version of an article that has been published in this journal.}
\placetextbox{0.5}{0.985}{Changes were made to this version by the publisher prior to publication.}
\placetextbox{0.5}{0.97}{The final version of record is available at \href{https://doi.org/10.1109/TII.2022.3228680}{https://doi.org/10.1109/TII.2022.3228680}}%
\placetextbox{0.5}{0.05}{Copyright (c) 2022 IEEE. Personal use is permitted.}
\placetextbox{0.5}{0.035}{For any other purposes, permission must be obtained from the IEEE by emailing pubs-permissions@ieee.org.}%

\title{A scalable framework for annotating photovoltaic cell defects in electroluminescence images}


 



\author{Urtzi Otamendi, Iñigo Martinez, Igor G. Olaizola, Marco Quartulli

\thanks{This publication resulted (in part) from the PROMISE (Advances in Photovoltaic Solar Energy Operation and Maintenance Research) project (KK2019/00088), which is financed by the ELKARTEK program of the Basque Government.}
\thanks{Urtzi Otamendi, Iñigo Martinez, Igor G. Olaizola and Marco Quartulli are with the Vicomtech Foundation, Basque Research and Technology Alliance (BRTA), Donostia-San Sebastián 20009, Spain (e-mail: uotamendi@vicomtech.org).}}

\markboth{Journal of Transactions on Industrial Informatics}%
{Otamendi \MakeLowercase{\textit{et al.}}: Deep learning-based defect extraction in PV cells}

\maketitle

\begin{abstract}
The correct functioning of photovoltaic (PV) cells is critical to ensuring the optimal performance of a solar plant. Anomaly detection techniques for PV cells can result in significant cost savings in operation and maintenance (O\&M). Recent research has focused on deep learning techniques for automatically detecting anomalies in Electroluminescence (EL) images. Automated anomaly annotations can improve current O\&M methodologies and help develop decision-making systems to extend the life-cycle of the PV cells and predict failures. This paper addresses the lack of anomaly segmentation annotations in the literature by proposing a combination of state-of-the-art data-driven techniques to create a Golden Standard benchmark. The proposed method stands out for (1) its adaptability to new PV cell types, (2) cost-efficient fine-tuning, and (3) leverage public datasets to generate advanced annotations. The methodology has been validated in the annotation of a widely used dataset, obtaining a reduction of the annotation cost by 60\%.
\end{abstract}

\IEEEpeerreviewmaketitle

\begin{IEEEkeywords}
photovoltaic cells, deep learning, anomaly segmentation, Golden Standard, electroluminescence, benchmark dataset, defect annotation
\end{IEEEkeywords}

\section{Introduction}\label{sec:introduction}

In recent years, the solar energy industry has emerged as one of the essential methods for renewable energy generation. The International Energy Agency (IEA) forecasts that solar PV installations will reach over 160 GW by 2022 \cite{IEA_forecast}. According to several studies \cite{IEA_road}, growth in solar energy generation has accelerated in recent years, with an increased capacity of more than 50\% during the last two years. Such expansion has brought up the financial and environmental benefits of implementing advanced operation \& maintenance (O\&M) processes \cite{ mustafa2020environmental}, and the research community has begun to develop artificial intelligence techniques to improve operational efficiency and predictive maintenance \cite{ hernandez2019review}.

The proper operation of photovoltaic cells (PV) is critical to ensuring the optimal performance of a solar plant. Damage to PV cells frequently occurs during manufacturing, and operators conduct extensive inspections to detect damaged cells. These failures may occur during plant installation or operation. Addressing this issue efficiently and preventing worst-case scenarios should lead to lower O\&M costs.

There are various techniques for monitoring solar panels, both during installation and operation. For instance, Electroluminescence (EL) imaging techniques detect defects in a non-intrusive and effective manner. This imaging method has emerged as a powerful tool for detecting subtle defects such as micro-cracks or cell degradation \cite{el_extr_intr}.
Detecting defects in the EL images of a solar plant is a time-consuming and laborious task that requires domain experts. 
Automated anomaly detection can be a breakthrough change in plant O\&M, reducing operational costs and increasing Overall Equipment Effectiveness (OEE). Therefore,  the research community has proposed artificial intelligence-based approaches such as deep learning models \cite{Deitsch2019, spataru2015quantifying}.

However, the information provided by anomaly detection approaches is insufficient to extract detailed information about the anomaly causing the improper behavior. In this sense, segmentation offers detailed information about anomalies, such as shape or location. Therefore the research community has started to focus on applying deep learning-based anomaly segmentation methodologies. Nevertheless, training these models requires segmentation annotations that are very expensive to obtain. Thus, efforts have focused on exploiting classification annotations using semi-supervised and weakly-supervised techniques \cite{mayr, rahman2020defects}. Despite promising results, mentioned approaches are not as reliable as supervised segmentation models \cite{minaee2021image}.

\begin{figure*}[t]
    \centering
    \includegraphics[width=0.82\linewidth]{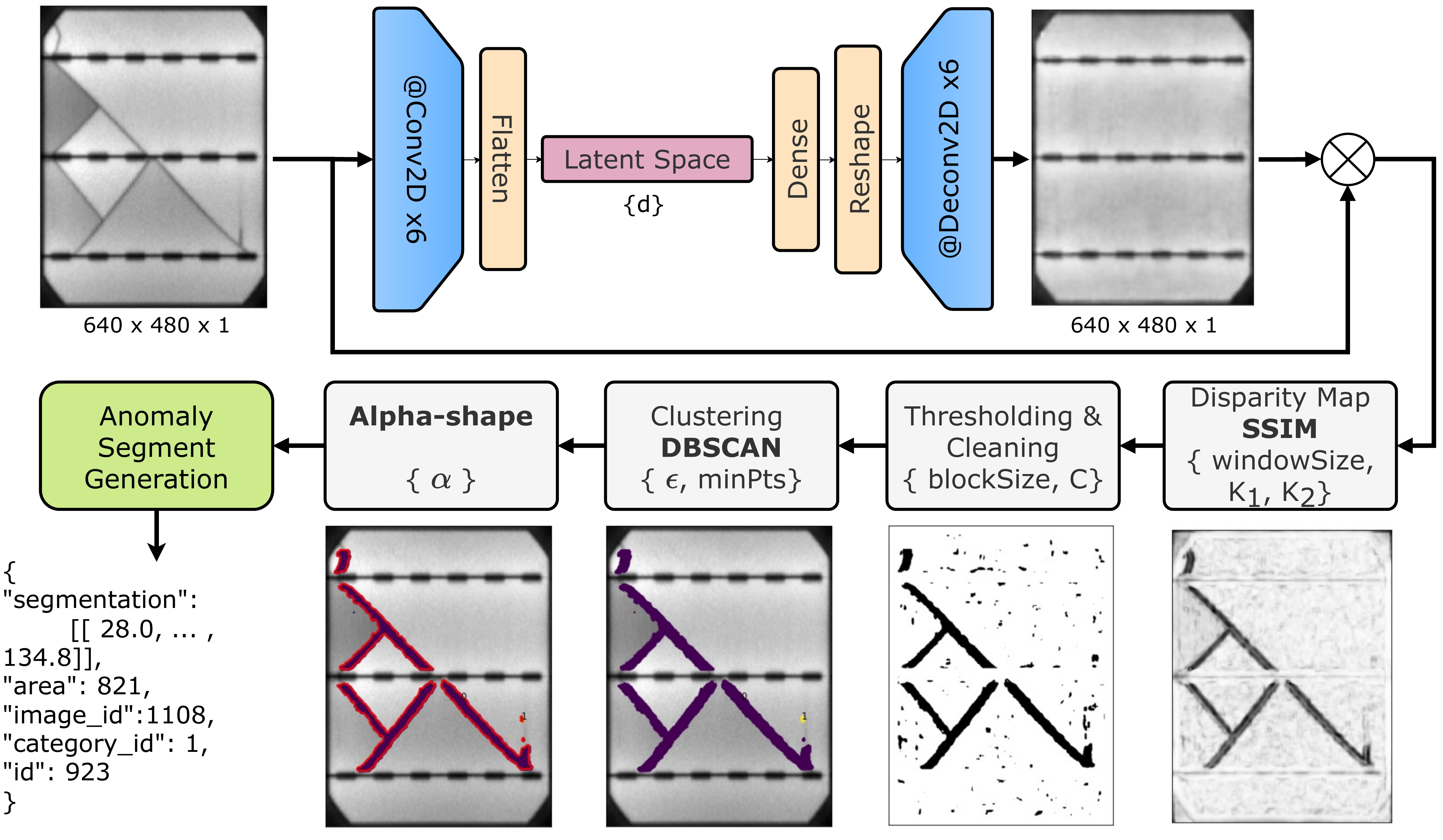}
    \caption{An overview of the proposed methodology for detecting and segmenting anomalies in electroluminescence images of PV cells. First, a weakly-supervised deep autoencoder generates an anomaly-free image from the original PV cell; Then, the SSIM \cite{SSIM} metric is used to calculate the disparity map between the original and generated images. Finally, the PV cell anomalies are detected and segmented using a pipeline of unsupervised techniques (thresholding \& cleaning + DBSCAN clustering \cite{ester1996density} + alpha-shape algorithm).}
    \label{fig:diagram}
\end{figure*}

Recent contributions have conducted experiments using supervised learning to train segmentation models. Nonetheless, these works are based on private datasets and do not publish segmentation annotations. The lack of a standard collection results in the impossibility of unifying the state-of-the-art and obtaining an accurate evaluation and measurement of the model performance, thereby reducing the field's evolution.

In this sense, a \textit{Gold Standard Corpora} (GSC) is a term used in natural language processing to denominate a standard collection. The creation of GSC is a time-consuming process usually performed by domain experts. In the event of the impossibility of manually annotating [Rebholz-Schuhmann et al., 2010] \textit{Silver Standard} approach is used. Standard silver annotations, usually generated automatically, will not be of the same standard quality as manual annotations but are very reliable. These annotations' quality distinguishes between the gold standard and uncontrolled automatic annotation.
This article proposes a data-driven methodology to establish a Silver Standard in the domain of Photovoltaic cells. The proposed method extracts in a cost-efficient manner anomaly segments from EL images, and it is adaptable to detect annotations across different PV cell types. This contribution is intended to provide a standard annotation of a large variety of cell types to constitute a benchmark for the state-of-the-art.

The proposed methodology combines state-of-the-art data-driven techniques to create the processing pipeline. This pipeline relies on a weakly-supervised deep-learning model and unsupervised clustering techniques. The deep-learning model is trained using non-defective images of one type of PV cells, allowing it to learn the structural distribution of non-defective cells and detect differences with the defective cells. The rest of the pipeline extracts the segments that differ from a structural distribution of a non-defective cell.

The data-driven methods learn from a large amount of data. The low variability of cell types facilitates the training of the models, yielding cost-effective training and obtaining high performance with a low number of samples. Thus, this characteristic results in high adaptability to new PV cell types.

In this sense, the presented method contributes to the PV plant O\&M process and the research community. On the one hand, due to the low diversity of cell types in a single solar power plant this approach can be used enhance the capacity of the solar plant to detect and analyze anomalies cost-efficiently. On the other hand, the segmentation data can be processed as annotations to provide the research community with a Silver Standard benchmark. Ultimately, a domain expert can review the dataset to raise the quality of the annotations, considerably reducing the cost of creating a \textit{Gold Standard}.


In summary, this paper proposes a data-driven methodology for extracting anomaly segments from EL images that is also cost-effectively adaptable to detect anomaly annotations across different PV cell types. The method combines state-of-the-art deep learning techniques to create a processing pipeline. These are the main contributions: (1) We propose a novel approach that combines a weakly-supervised deep-learning model with unsupervised clustering techniques to address the field's lack of annotations. (2) We present an algorithm for producing \textit{Gold Standard} annotations of large PV cells datasets as a benchmark (3) This methodology reduces the cost of anomaly segment annotation by $2.71$, from $19.9$ seconds to $7.337$ seconds. (4) We apply the proposed contribution in public and private PV cell datasets with a wide range of cell types as part of the adaptability validation phase.

\section{Related work}\label{sec:relatedwork}

The electroluminescence technique is based on the optical phenomenon in which the PV cell material emits light in response to an electric current. Admittedly, this current stimulates the PV module and reverses its operation \cite{kim2005electroluminescence}. Consequently, PV cells start to emit light at a wavelength peak of $1150 nm$, revealing the possible defects in a PV cell \cite{fuyuki2005photographic}. The light is captured by a special camera, typically equipped with special optical filters, obtaining an EL image of the PV cell.
This technique can detect various PV cell anomalies that would otherwise go undetected by other methods. The EL procedure is preferred for detecting structural defects, intrinsic defects (i.e., fingerprint marks), and extrinsic defects (e.g., micro-cracks, cell degradation, corrosion, electrically isolated parts), according to \cite{reviewPana}. EL is the only method to detect non-electrical active cracks at the cell level, thus positioning EL as the preferred method to perform such anomaly detection tasks. 
Additionally, Ruizhen Yang \textit{et al.} \cite{Yang2020ElectromagneticIH} proposed an approach to combine the electrothermography (ET) and electroluminescence (EL) detection effects of defects. This work studies electromagnetic induction (EMI) and image fusion, demonstrating how EMI can significantly enhance ET and EL's capacity for defect detection by combining and complementing the two wavelength detection data.

The EL procedure is regarded as one of the least invasive and least expensive methods. The resulting images are of high quality, which facilitates the detection of subtle anomalies. However, capturing EL images can be difficult due to the required capture conditions \cite{mochizuki2016solar, fuyuki2005photographic}: the panels must be cold and in complete darkness to avoid residual radiation emission. Due to these difficulties, the availability of correctly annotated public EL datasets is limited.

Anomaly detection research works regarding EL imagery have primarily been developed using the dataset \textbf{ELPV} published by \textit{Sergiu Deitsch et al.} \cite{ Buerhop2018,Deitsch2019, Deitsch2021}. The dataset contains $2624$ images with a resolution of $300\times300$, captured from 44 PV modules: 26 poly-crystalline and 18 mono-crystalline. Although there are several types of anomalies (e.g., micro-cracks, electrically insulated, disconnected cells, and degradation), only four-class annotations are provided. The annotations classify the PV cell according to the defect likelihood (0.0, 0.33, 0.66, 1.0).

In \cite{karimi2019automated} \textit{Ahmad Maroof Karimi et al.} propose an automated pipeline for PV module EL image processing. The authors use a dataset of $5400$ PV cell images captured from three different PV modules. The images were classified as cracked, corroded, or non-defective, but since the authors did not publish the dataset, no further work was performed.

Recently more attempts have been made to address anomaly detection via EL imagery \cite{tang2020deep, akram2019cnn}. The majority of these approaches, however, either use the previously mentioned dataset (ELPV) or rely on undisclosed private data. Furthermore, these approaches use non-supervised learning techniques, such as semi-supervised or weakly-supervised learning, to segment anomalies: \textit{Mayr et al.}\cite{mayr} proposed a weakly supervised strategy to perform anomaly segmentation in electroluminescence imagery. This method uses a modified version of ResNet-50 to extract segmentation via the network's activation maps. The authors apply an \textbf{\textit{$L_{p}$ }} normalization to aggregate the activation maps into single scores for classification. \textit{Rahman et al.}\cite{rahman2020defects} proposed a multi-attention network to efficiently extract the most important features or EL imagery. This method can segment complex anomalies but must be trained on annotated data. \textit{Pierdica et al.}\cite{pierdicca2020automatic} proposed an automatic anomaly segmentation in infrared imagery based on Mask R-CNN architecture. In this work, the authors compare the approach with the three state-of-the-art anomaly classification networks: UNet, FPNet, and LinkNet.

From this review we conclude that there is a lack of segmentation annotations for PV cell electroluminescence imagery. This lack of annotations implies an additional difficulty for domain expert users to develop anomaly segmentation approaches. As a result, most processes tend to use semi-supervised and weakly-supervised techniques to create machine learning models, but these approaches present a lack of generalization concerning variation in cell or anomaly types.

\section{Methodology}\label{sec:Methodology}

This section describes the proposed method for detecting and segmenting anomalies across PV cell types. The methodology is based on state-of-the-art weakly supervised and unsupervised techniques and consists of five processing steps divided into two parts. The first one employs a deep-learning model to generate an anomaly-free image from the original PV cell electroluminescence capture (see fig. \ref{fig:diagram} upper row). The second part uses unsupervised machine learning algorithms to process the \textit{Structural Similarity Index Metric} (SSIM) \cite{SSIM} disparity map between the original and generated images (see fig. \ref{fig:diagram} lower row).

The processing pipeline first receives a defective PV cell image. The deep autoencoder then infers the \textit{non-defective} version of the original image. Following a supervised learning schema, the autoencoder is trained on non-defective PV cell imagery. Given the reduced number of parameters in the deep autoencoder, this training process is computationally inexpensive, allowing the model to adapt to new PV cell-type data efficiently.
Following that, the original image and the \textit{non-defective} generated image proceed to the unsupervised processing stage, in which a disparity map is first computed using both images. This disparity map is subsequently binarized using a thresholding and cleaning algorithm. Then, the coordinates of the different pixels are extracted from the resulting image and later clustered using the  \textit{Density-based spatial clustering of applications with noise} (DBSCAN) \cite{ester1996density} density clustering algorithm. Finally, the alpha-shape algorithm is used to compute the contour of the clustered points. These contours are ultimately exported as an anomaly segment annotation. All the aforementioned steps use pre-defined parameters that an operator can fine-tune to suit the methodology's performance to various PV cell types.

\subsection{Deep Learning model}\label{sec:DeepLearningModel}

The deep learning model must learn the distribution of non-defective PV cell images to detect defective areas as outliers from such non-defective distribution. Based on the segmentation method proposed by \textit{Otamendi et al.} \cite{otamendi2021segmentation}, a deep learning technique was considered to infer higher-level (segmentation) annotations from the classification of the images. Their paper discusses the shortcomings of the proposed method, especially those related to the lack of data. Some observations suggest that the performance is decreased due to the variation of PV cells. Nonetheless, the variation data in this paper is not a concern since our method stands out for its adaptability to new PV cell types. 

In this case, the objective is to fine-tune the base model for significant variations of PV cell types. Therefore, whenever a new dataset needs to be analyzed and segmented, the model is adjusted to learn the distribution of the new PV cell types. The low variability training set facilitates cost-effective training that can be performed using small datasets. In this sense, when processing a defective cell image ($I$), the auto-encoder will generate a non-defective representation ($\hat{I}$) to highlight the defective areas appropriately. As a result, this process addresses anomaly segmentation as an outlier detection task, where a defective point is a data point that differs from the observed data distribution.

\begin{table}[b]
    \vspace{-2em}
    \centering
     \caption{Autoencoder topology for the segmentation phase. In total, the proposed model has $15,417,913$ trainable neurons.}
     \resizebox{0.8\linewidth}{!}{%
    \begin{tabular}{ccccc}
        \toprule
        Layer & Output Shape & Filters & Kernel & Stride\\ \midrule
        Input & $640\times480\times1$ \\
        Conv2D & $320\times240\times32$ & $32$ & $2\times2$ & $2\times2$ \\
        Conv2D & $160\times120\times16$ & $16$ & $2\times2$ & $2\times2$ \\
        Conv2D & $160\times120\times8$ & $8$ & $4\times4$ & $1\times1$ \\
        Conv2D & $80\times60\times16$ & $16$ & $2\times2$ & $2\times2$ \\
        Conv2D & $80\times60\times8$& $8$ & $4\times4$ & $1\times1$ \\
        Conv2D & $80\times60\times16$& $16$ & $4\times4$ & $1\times1$ \\
        Conv2D & $80\times60\times8$ & $8$ & $4\times4$ & $1\times1$ \\
        Flatten & $38400$ \\
        Dense & $200$ \\ 
        Dense & $38400$ \\ 
        Deconv2D & $80\times60\times8$ & $8$ & $4\times4$ & $1\times1$ \\
        Deconv2D & $80\times60\times16$& $16$ & $4\times4$ & $1\times1$ \\
        Deconv2D & $80\times60\times8$& $8$ & $4\times4$ & $1\times1$ \\
        Deconv2D & $160\times120\times16$ & $16$ & $2\times2$ & $2\times2$ \\
        Deconv2D & $160\times120\times8$ & $8$ & $4\times4$ & $1\times1$ \\
        Deconv2D & $320\times480\times16$ & $16$ & $2\times2$ & $2\times2$ \\
        Deconv2D & $640\times480\times1$ & $1$ & $2\times2$ & $2\times2$ \\
        Output & $640\times480\times1$  \\
        \bottomrule
    \end{tabular}
    }
    \label{table:autoencoder}
\end{table}

According to recent studies \cite{peloso2010observations, fuyuki2005photographic}, EL imagery cameras usually have a resolution of $640\times480$ pixels. Thus, the input layer of the model is set to the same size. The most promising topology is chosen for the convolutional layers after a phase of experimentation and analysis in which different autoencoder topologies were trained using a variety of PV cell types.

A relatively small number of convolutional layers has proven to be the most promising approach for propagating key features of PV cells. The distinguishing features of the PV cell technologies are slight, so a large number of convolutions would cause the vanishing of these features. On the other hand, a short number would reduce the ability to target and extract these key features. Therefore, the selected topology sits in the middle ground, composed of $14$ convolutional and deconvolutional layers. The last layer of the model uses sigmoidal activation, while the rest uses Leaky RelU activation.

Each layer is configured by different filters, kernel size, and stride, as seen in Table \ref{table:autoencoder}. The stride controls the way the filter convolves around the input. Large stride values reduce the spatial dimension of the input and help receptive fields to overlap less. In this case, we want to avoid a significant loss of information. Thus we will use small stride values. 

The latent space layer, which connects the encoder and decoder, is used to adjust the model performance. Considering this work's objective and use case, a relatively small latent space dimension ($200$) is recommended, which significantly reduces the size of the neural network, making it more lightweight. Consequently, training costs are reduced, facilitating the adaptability to new PV cell types.

According to \textit{Otamendi et al.} \cite{otamendi2021segmentation}, the \textit{Structural Similarity Index Metric} (SSIM) \cite{SSIM} was selected as loss function of the deep learning model. The SSIM is a perceptual metric that has demonstrated excellent performance in measuring the similarity of two images. In contrast to pixel-wise independent loss functions like L2, SSIM computes the similarity using image illumination, contrast, and structure. SSIM measures the similarity of two images with a value of $[-1,1]$, where $-1$ represents completely different images, and $1$ represents completely similar images. Therefore, given that the objective of training a model is to minimize the loss function and our objective is to maximize SSIM, our loss function will be the minimization of the negative SSIM.

\begin{equation} \label{eq:ssim} 
SSIM(x,y)=\frac{(2\mu_{x}\mu_{y}+(k_{1}L)^{2})+ (2\sigma_{xy}+(k_{2}L)^{2})}{(\mu_{x}^{2}+\mu_{y}^{2}+(k_{1}L)^{2})(\sigma_{x}^{2}+\sigma_{y}^{2}+(k_{2}L)^{2})}
\end{equation} 

In this work, the size of the Gaussian filter for SSIM is set to 7. As a result, the SSIM metric will receive two sliding-window inputs, $x$ from the original image ($I$) and $y$ from generated image ($\hat{I}$), both with a size of $7x7$. In addition, as in the original paper, the rest of constants will be set as follows: $k_{1}=0.001$, $k_{2}=0.03$ and $L$ (dynamic range of the pixel values) in this case will be $L=255$.

\subsection{Disparity Map}\label{sec:disparitymap}

The next step is to compute the disparity map, i.e., the difference between the original and generated images. The disparity map reveals even the slightest variations and provides enough information to detect the areas outside the distribution of a non-defective PV cell. This map is obtained using the previously mentioned SSIM metric, where a pixel disparity is measured between -1 and 1 (see fig. \ref{fig:threshold_compare} a).

\subsection{Thresholding}\label{sec:Thresholding}

Anomalies tend to have a continuous disparity distribution throughout the segment, e.g., partial breakage or micro breakage. However, noise in the original image can cause structural differences propagated through SSIM into this stage. Therefore, a thresholding process must be applied to focus on the areas where the disparity is more accentuated and prominent. For this purpose, different thresholding techniques were tested to obtain the most accurate results.

Thresholding is a segmentation technique that separates the foreground from the background, resulting in a binary image. The most basic method compares the intensity of each pixel to a fixed constant T to discriminate the segments in the image. There are several methods for determining the value of the T constant automatically, such as Otsu's thresholding \cite{otsu1979threshold}, a popular clustering-based technique. When the image has a bi-modal distribution, this model performs well. However, the algorithm has difficulty determining the foreground and background in noisy images. This method employs a global thresholding technique, so the segmentation is performed using the same T. On the other hand, the adaptive thresholding approach can provide more detailed segment extraction by computing different T values for local regions.

\begin{figure}[t]
    \vspace{0em}
    \centering
    \includegraphics[width=\linewidth]{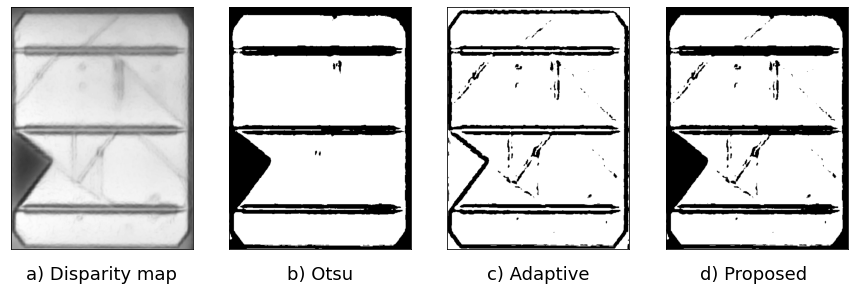}
    \caption{Thresholding method comparison in a representative PV cell electroluminescence image.}
    \label{fig:threshold_compare}
\end{figure}

To exemplify the operation and performance of the thresholding process, we will use a representative image of a damaged photovoltaic cell (see fig. \ref{fig:threshold_compare}a). First, we used a method based on the arithmetic mean as the adaptive thresholding (see fig. \ref{fig:threshold_compare}c). This method calculates the average intensity of the neighboring pixels within the local region, which in this case is defined as $41$ pixels. This value must be large enough to cover both the foreground and background areas. On the other hand, we compared the performance to Otsu's advanced thresholding method \cite{otsu} (see fig. \ref{fig:threshold_compare}b).

Otsu's method detects areas with significant disparity, as shown in figure \ref{fig:threshold_compare}, but it performs poorly in detecting variations in low disparity areas. On the other hand, the adaptive method detects disparity gradients since it uses local regions. As a result, it detects anomalies with a low disparity value and large disparity gradient compared to its neighbors. However, it is less accurate in slightly varying areas with a high disparity for the overall image. Therefore, we propose combining both methods to leverage each other's strengths while compensating for their weaknesses. As shown in Fig. \ref{fig:threshold_compare}d, the image represents areas with high local gradient variation and global disparity. 

\subsection{Noise Cleaning}\label{sec:Cleaning}

The creation of the disparity map may produce noise, particularly in the image's corners and near the busbars, primarily due to the imprecise framing of PV cells during the image capture process. 
The image's border and busbars do not emit light during the electroluminescence process, so the structural difference near these areas is significant. As a result, we included an additional step to detect the edges and busbars and clean up any potential noise in the surrounding area.

\subsection{Clustering}\label{sec:Clustering}

Following the detection and cleaning of disparity points, the next step is to cluster these points. Due to the lack of segmentation annotations, the method uses an unsupervised clustering algorithm based on point density for this task. Based on state-of-the-art clustering techniques, the \textit{Density-based spatial clustering of applications with noise} (DBSCAN) \cite{ester1996density} algorithm was found as the most suitable for the intended application. 

\begin{figure}[t]
    \centering
    \includegraphics[width=1.0\linewidth]{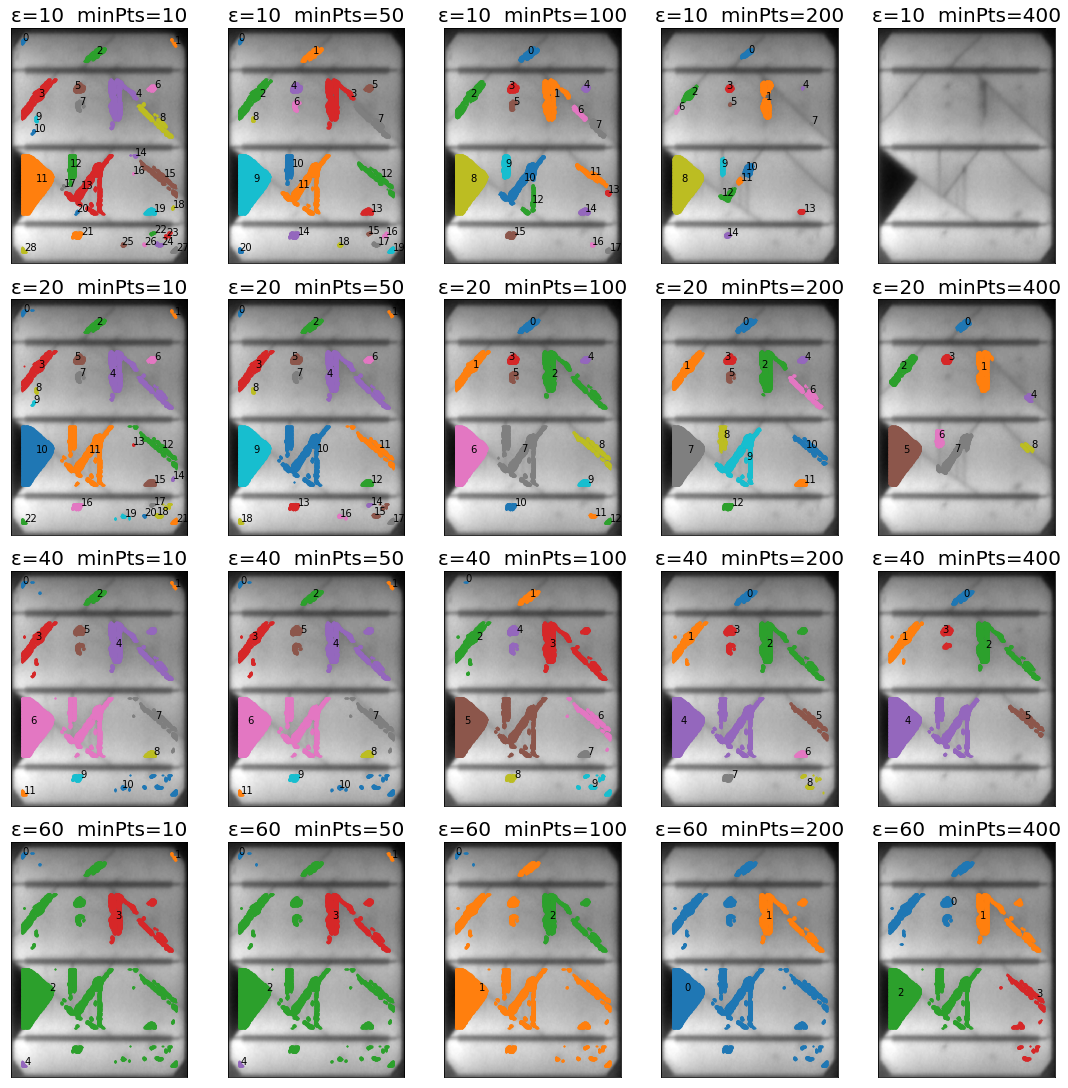}
    \caption{ DBSCAN clustering method performance comparison using different values of $\epsilon$ and \textit{minPts}.}
    \label{fig:cluster_compare}
\end{figure}

DBSCAN can extract structured patterns in data. It uses two parameters, \textit{$\epsilon$} and \textit{minimum points of a cluster} (minPts), to classify points as core, border, and outlier points. Core points have at least \textit{minPts} within distance $<\epsilon$. Border points are reachable from a core point ($<\epsilon$), but it has less than \textit{minPts} in the surrounding area. The outlier point is not reachable from any core point. Low $\epsilon$ values will result in many points being classified as outliers, while a large value will cause the noise to be classified within a cluster. Moreover, large \textit{minPts} values are useful for data sets with noise. Based on a series of experiments (see fig. \ref{fig:cluster_compare}), the optimal value for $\epsilon$ was determined to be near $30$ and for \textit{minPts} near $100$.

\subsection{Alpha-shape}\label{sec:Alpha-shape}

Geometrical structures are then extracted from the resulting clusters by computing the hull of each cluster's point cloud. We addressed this task as a convex hull problem: the intersection of all convex sets containing a given subset of points. However, this technique struggles to generate representative hulls due to the complexity of the PV cell anomalies, which have many concave shapes. Therefore, a generalization concept of a convex hull called alpha-shape \cite{edelsbrunner1983shape} was preferred. This technique is related to the Delaunay triangulation sub-graph of the point set: two points are connected whenever there is no other point within a generalized disk of radius $1/\alpha$. If $\alpha$ is $0$, the radius is replaced by a closed half-plane, and the resulting alpha-shape is an ordinary convex hull. 

\begin{figure}[!h]
    \centering
    \includegraphics[width=0.9\linewidth]{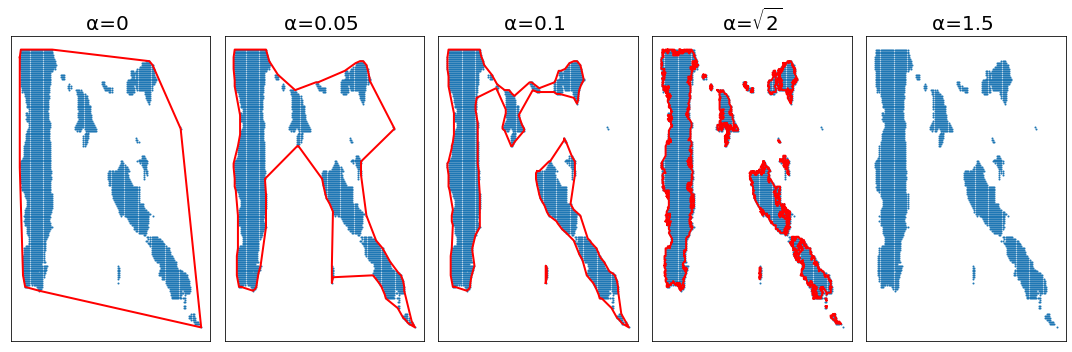}
    \caption{ Alpha-shape algorithm comparison for different values of $\alpha$.}
    \label{fig:alpha_compare}
\end{figure}

The performance of the alpha-shape algorithm can be fine-tuned using the appropriate $\alpha$ value. The radius of the generalized disk is defined as $1/\alpha$. Considering that a point within a cluster represents a pixel of the image, the distance between two neighbor points can only be $1$ or $\sqrt{2}$. Therefore, when the diameter exceeds the maximum distance ($\sqrt{2}$), the algorithm will not find a generalized disk connecting two points. This means that the maximum radius for the algorithm to work is $r=\sqrt{2}/2=1/\alpha$, leading to a maximum value of $\sqrt{2}$ for $\alpha$ (see fig. \ref{fig:alpha_compare}).

Finally, the resulting alpha-shape coordinates are converted to a high-label segmentation annotation format. The annotation generation process follows the standard used on principal segmentation challenges such as COCO and PascalVOC.

\section{Results}\label{sec:results}

 This section discusses the performed experiments and results, which validate this paper's contribution. It will explain the experimentation performed on a public dataset, detailing the training and pipeline tuning process and the time cost. Afterward, we will compare the annotation cost of the proposed pipeline with the manual process. Finally, the methodology has been tested on other types of cells to validate the adaptability of the approach. Experiments were performed using a GPU Tesla T4 with 16Gb of memory.

\subsection{Dataset}\label{sec:dataset}


The model was trained using the public ELPV dataset \cite{ Buerhop2018, Deitsch2019, Deitsch2021}, a widely used dataset in the literature on PV cell anomaly detection. This dataset was also selected to validate the methodology as it is the nearest dataset to a standard state-of-the-art benchmark.

The dataset is annotated using a four-class classification according to the defect likelihood of each PV cell: ($0.0/0.33/0.66/1.0$). The training of the deep autoencoder only needs non-defective images, so only the subset classified with a likelihood of 0 was selected. The distribution of the non-defective images is as follows: $1508$ images in total, of which $588$ are mono-crystalline PV cells, and the remaining $920$ are poly-crystalline. The images have a resolution of $300\times300$ pixels and are in 8-bit grayscale, and they were resized to fit the shape of the autoencoder input layer ($640\times480\times1$). Additionally, images were normalized to obtain a clearer contrast and improve the model's training. 

\begin{figure}[t]
\vspace{1em}
\centering
\subfloat{
\includegraphics[width=0.18\columnwidth]{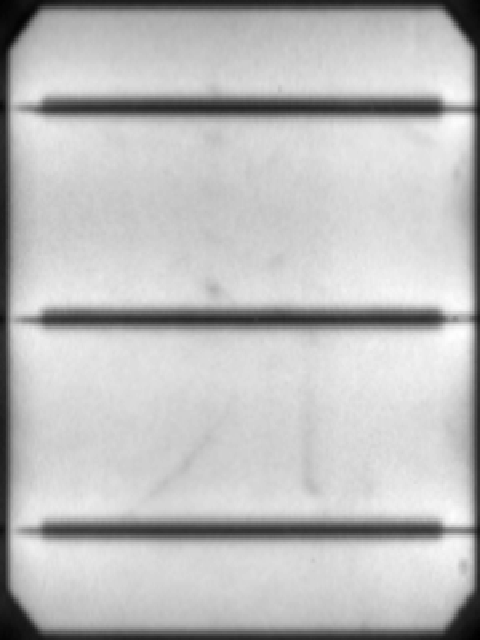}
\HHquad
\includegraphics[width=0.18\columnwidth]{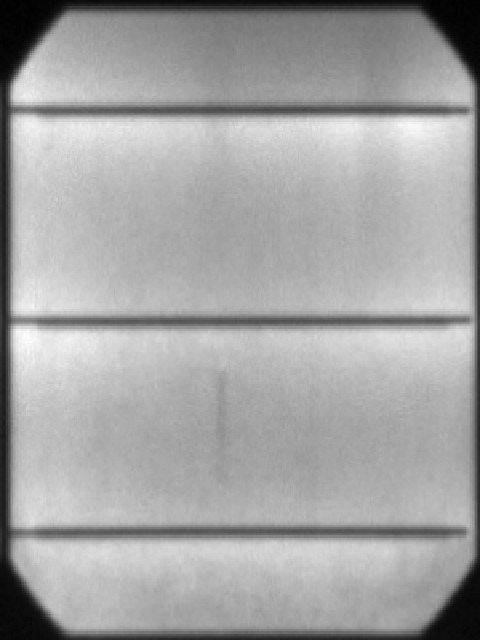}
\HHquad
\includegraphics[width=0.18\columnwidth]{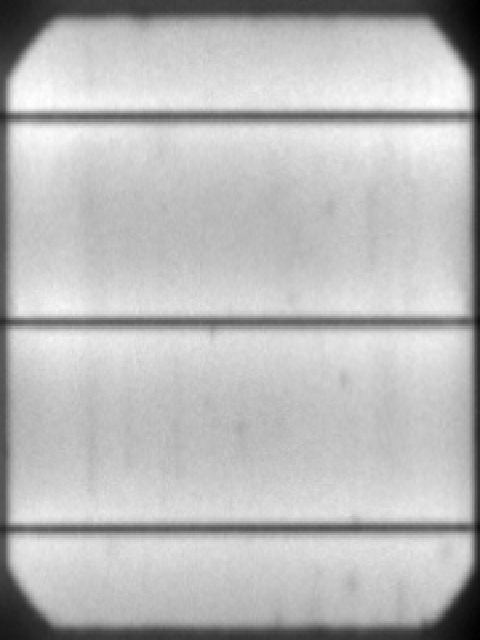}
\HHquad
\includegraphics[width=0.18\columnwidth]{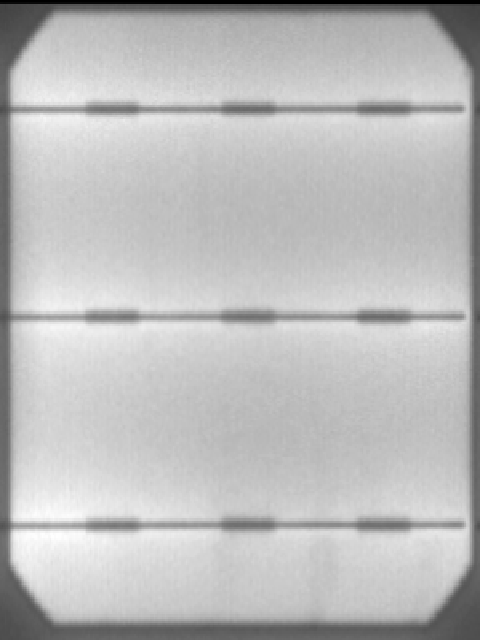}
\HHquad
\includegraphics[width=0.18\columnwidth]{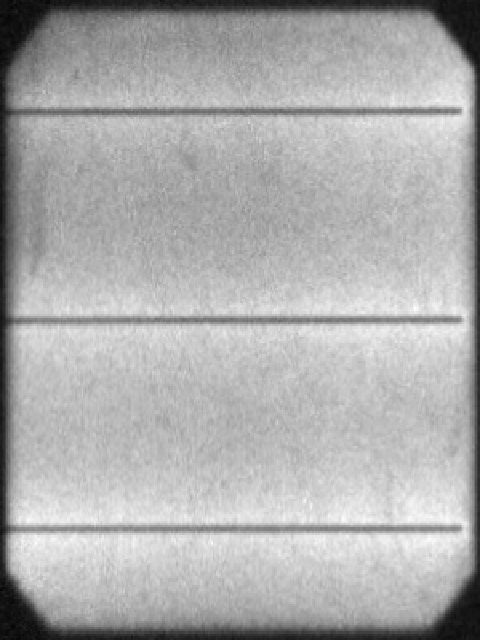}
\HHquad
}
\caption{Example of similar PV Cells types selected for the deep learning model training. }
\label{fig:example_similar_types}
\end{figure}

The proposed methodology learns from the structural distribution of non-defective PV cell images. Unfortunately, in the case of poly-crystalline cells, the structural distribution is not correctly preserved since the type of material generates non-uniform patches in the images \cite{Deitsch2019}. Therefore, the autoencoder model cannot learn the distribution of non-defective cells, as there is no uniform distribution.
Therefore, for the experimentation, only mono-crystalline PV cells were chosen. These cells have structural differences, primarily due to the variation in the number of busbars, so the flexibility of the model can be verified (see Fig. \ref{fig:example_similar_types}).

The initially selected subset of $588$ images was augmented to $2352$ images employing simple transformations, such as flip and rotation. These transformations enlarge the size and diversity of the dataset while avoiding excessive modifications in the original non-defective images. Considering that a balanced dataset usually helps to perform unbiased training, the augmentation task was done so that it balances and enriches the distribution of the data. In addition, due to the sigmoidal activation, the model output is an image with values in the $[0,1]$. Therefore, to perform the training adequately, images were re-scaled from values ranging from $[0,255]$ to $[0,1]$. Finally, the dataset was split into $80\%$ training set and $20\%$ validation set.

\subsection{Pipeline tuning}\label{sec:Pipeline_tuning}


This section presents the adjusted model parameters for the experiments performed with the ELPV dataset. These initial values can be used as a reference and be modified to adapt the performance to new data. For the training task, we used Adam optimization with a learning rate of $0.003$, a $\alpha$ value of $0.025$ for the Leaky RelU, and a batch size of $16$. The training was performed during $200$ epochs, with a validation phase every $5$ epochs. Additionally, we included early stoppage with the patience of $10$ epochs to avoid unnecessary training. This process took $20.5$ minutes to complete.

\begin{figure}[b]
    \centering
    \includegraphics[width=0.9\linewidth]{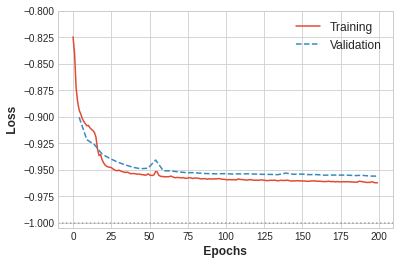}
    \caption{ The training process of the autoencoder on the ELPV subset: training and validation loss at each epoch (\textit{EST: 20.5 min}). }
    \label{fig:training}
\end{figure}

The autoencoder model uses the negative SSIM as the loss function, which returns a value in the range of $[-1,1]$ where $-1$ equals $SSIM=1$. As shown in Fig. \ref{fig:training}, the training and validation loss were very close during the entire process, and the training loss converged at $-0.965$. Finally, the model was validated using a small sample of $100$ images and the SSIM metric to compute the accuracy of the autoencoder. The model obtained a mean accuracy of $0.943218$, a median value of $0.9535$, and the accuracy range was $[0.9235,0.9752]$.

\newcommand\y{0.115}
\begin{figure*}[!h]
\centering
    \subfloat[][Original]{\includegraphics[width=\y\linewidth]{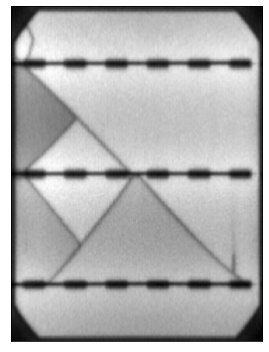}} \Hquad
    \subfloat[][Autoencoded]{\includegraphics[width=\y\linewidth]{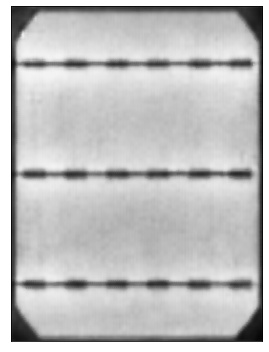}}\Hquad  
    \subfloat[][Disparity map]{\includegraphics[width=\y\linewidth]{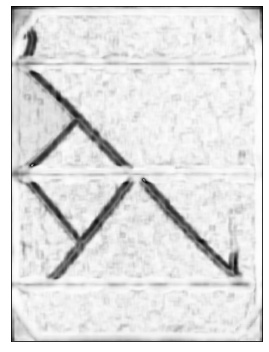}} \Hquad
    \subfloat[][Thresholded]{\includegraphics[width=\y\linewidth]{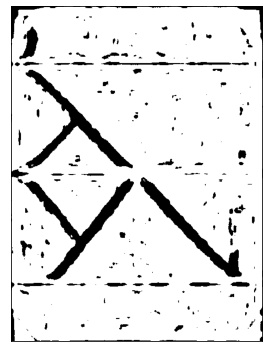}} \Hquad
    \subfloat[][Cleaned]{\includegraphics[width=\y\linewidth]{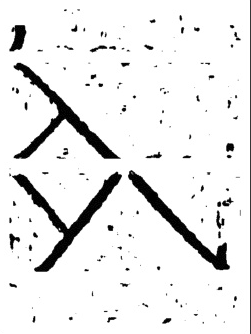}}\Hquad
    \subfloat[][Clusters]{\includegraphics[width=\y\linewidth]{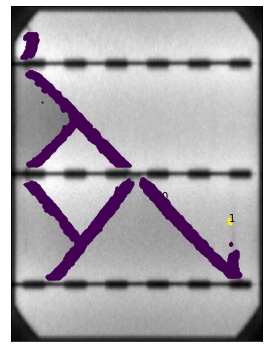}}\Hquad
    \subfloat[][Alpha-shape]{\includegraphics[width=\y\linewidth]{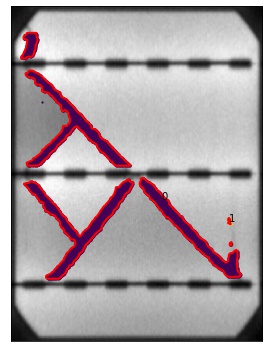}}\Hquad
    \subfloat[][Final]{\includegraphics[width=\y\linewidth]{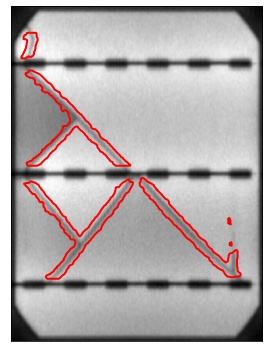}}
\caption{ Illustration of the proposed model's performance. Each image represents the output of each step: \textbf{b)} deep learning autoencoder, \textbf{c)} Disparity map via SSIM, \textbf{d)} Thresholding via the proposed method, \textbf{e)} Noise cleaning, \textbf{f)} Point clustering via DBSCAN, \textbf{g)} Convex Hull via Alpha-shape.\textbf{ More examples can be found in the appendix \ref{apen:a}} }
\label{fig:step_by_step}
\end{figure*}


Afterward, performing a few tests that took $12$ minutes, the unsupervised models' parameters were tuned to adjust the performance to the PV cell type. An SSIM window size of $11$, $K1=0.001$, and $K2=0.05$ was used to compute the disparity map. An adaptive block size of $61$ and $C=10$ was selected in the thresholding process. Finally the clustering process used $\epsilon=10$ and $minPts=100$ values. Fig. \ref{fig:step_by_step} shows a step-by-step example of the performance of the proposed methodology, which can extract the anomaly segment from the PV cell (see appendix \ref{apen:a} for more examples).

\subsection{Cost Comparison}\label{sec:cost_comparison}

After performing the adaptation and tuning of the pipeline, we applied the proposed methodology to generate annotations for the dataset. In this process, $486$ mono-crystalline defective cells have been annotated. Once this process has ended, we have obtained the Silver Standard annotations for the ELPV dataset.

\begin{figure}[b]
    \centering
    \includegraphics[width=0.9\linewidth]{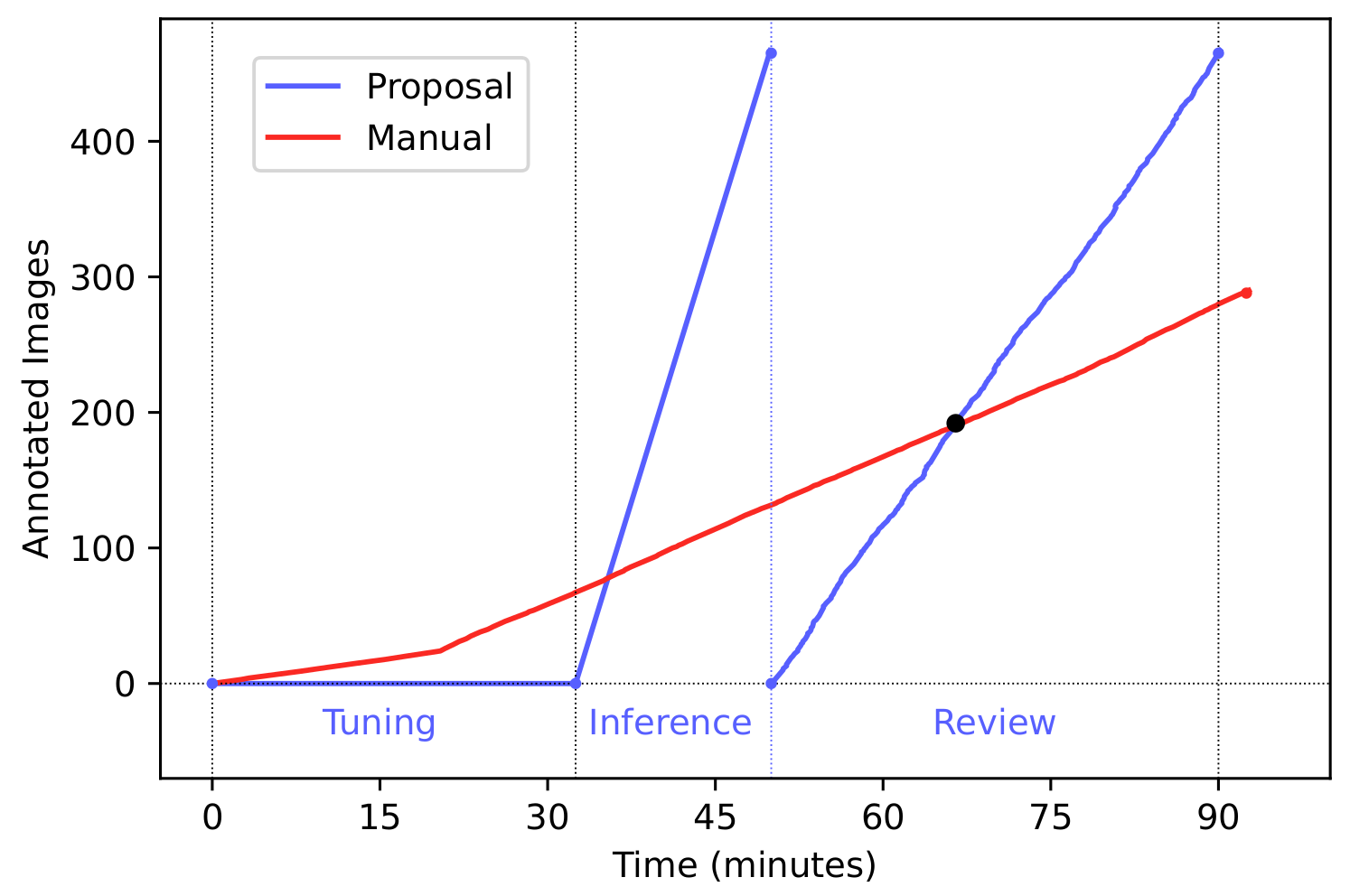}
    \caption{ Comparison of the annotation cost of the proposed approach and the manual annotation of a domain expert. The proposed performance is divided into three phases: tuning, inference, and manual revision.}
    \label{fig:compar}
\end{figure}

The entire generation process has taken $50$ minutes to complete, $32$ for the model training and pipeline tuning, and $17$ for the annotation inference, $2.2$ seconds of running time of the proposed algorithm per image. Following this process, we have performed an extra manual annotation review step to elevate the annotations' quality from a Silver Standard to a Golden. This step has taken $43$ minutes, $5.3$ seconds of review time per image.
In order to compare the cost of manual and automatic processing, a domain expert has annotated the same dataset in the same period of time as the process described above using the annotation tool by M. Tkachenko \textit{et al.} \cite{Label_Studio}. As seen in figure \ref{fig:compar}, the expert has managed to annotate $271$ images in $90$ minutes, $19.9$ seconds of annotation time per image.

The results demonstrate that using the proposed methodology is significantly more efficient for generating golden standard annotations than manually annotating the images. Moreover, the adaptation and tuning phase is performed only once, so the more images there are, the lower the cost per image. Therefore, in this experiment using the proposed approach, the total cost per image ($t_{image}$) is $11.7$ seconds (see eq. \ref{eq:time}), where $t_{inference}$ is $2.237$ seconds,  $t_{revision}$ is $5.3$ seconds, $t_{tuning}$ is $1950$ seconds, and $n_{images}$ is $468$ images.

\begin{equation}
\centering
t_{image} = t_{inference} + t_{revision} + \frac{t_{tuning}}{n_{images}} 
\label{eq:time}
\end{equation}

\subsection{Adaptability}\label{sec:validation}

\begin{figure}[!ht]
\centering
    \subfloat{\includegraphics[width=.25\linewidth]{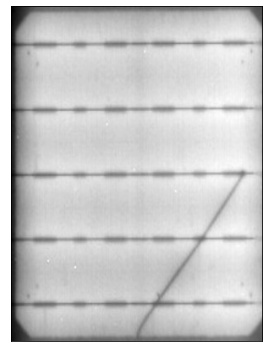}} \quad \quad
    \subfloat{\includegraphics[width=.25\linewidth]{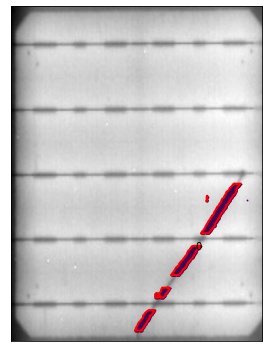}} \\
\caption{ Example of the behavior of the pipeline performed on a private dataset. This dataset contains images of mono-crystalline PV cells of 5 busbars. More examples can be found in the appendix \ref{apen:a}. }
\label{fig:private_performance}
\end{figure}

To validate the adaptability of the methodology, we used a private dataset composed of different PV cell types. For this experiment, we selected mono-crystalline cells of 5 busbars since they are the more challenging cells compared to the training. An initial subset of $1620$ images was later augmented to $6480$ using the same simple transformations mentioned in Section \ref{sec:dataset}. The training lasted for $164$ epochs using the previously mentioned parameter configuration. 
As shown in Figs. \ref{fig:private_performance} and \ref{fig:appendix_a}, a satisfactory result was obtained in both experiments, which demonstrates that the proposed methodology can process different PV cell types.

\section{Conclusions}\label{sec:conclusions}


This paper proposes a methodology to address the lack of PV cell anomaly segmentation annotations in the literature. The methodology combines data-driven techniques that efficiently and cost-effectively adapt to extract anomaly segments annotations from EL images and across various types of PV cells. Modern deep learning techniques are combined in the strategy to build a processing pipeline. The provided approach generates annotations of large PV cells datasets to constitute a benchmark in the field. 

Additionally, a domain expert can review these annotations to raise the quality and create a \textit{Gold Standard}. The proposed processing pipelines combine a weakly-supervised deep-learning model and unsupervised clustering that can be cost-effectively fine-tuned, even with a small dataset. Due to its adaptability, this methodology accurately processes PV cells with different structural features (see sec. \ref{sec:validation}).

In the experimental section, we validated the model's performance on ELPV public datasets and compared the temporal cost of annotating anomalous segments using the proposed methodology versus manually by an expert. According to the results, the automatic generation and manual review processes are more efficient than manual annotation. The inference ($2.2$) and review ($5.3$) takes $7.5$ seconds per image, whereas the manual annotation takes $19.9$ seconds (see sec. \ref{sec:cost_comparison}). Thus, the proposed methodology reduces $2.71$ times the cost of annotation. Although the methodology requires a tuning process once per dataset, we have shown that this can be done efficiently (see sec. \ref{sec:Pipeline_tuning}). Furthermore, to verify the adaptability of the proposed approach, we processed a private PV cell dataset with various cell types (see sec. \ref{sec:validation}).

This work can be used to establish a benchmark for the research community to develop and evaluate their segmentation models. Furthermore, high-level annotations will enhance the ability to train supervised models, thereby significantly increasing the accuracy and performance.
Regarding future work, the proposed algorithm accuracy could be improved by novel semi-supervised segmentation methods \cite{NEURIPS2021_b98249b3} that have been successfully applied for other domains, such as video semantic recognition \cite{7859408} or human activity recognition \cite{8767027}. Any improvement in accuracy would translate to a reduction in manual revision time.

\appendices
\section{Segmentation results}\label{apen:a}
The performance of the proposed methodology on mono-crystalline PV cells from the public ELPV dataset \cite{ Buerhop2018, Deitsch2019, Deitsch2021} is shown in this appendix. The segmentation regions detected by the proposed model are highlighted in red. Figure \ref{fig:appendix_a} contain 36 images in total, which are some of the most representative images within the dataset. Different cell types and anomaly classes were selected to validate the proposed methodology's performance in a variety of cases.

\newcommand\x{0.12}
\begin{figure*}[!ht]
\centering
\includegraphics[width=\x\linewidth]{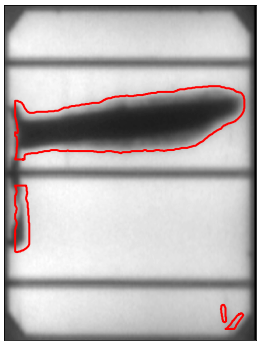}
\includegraphics[width=\x\linewidth]{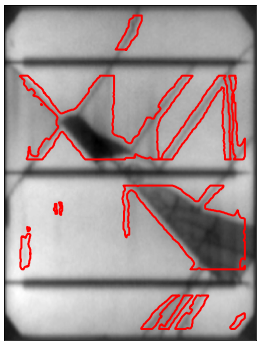}
\includegraphics[width=\x\linewidth]{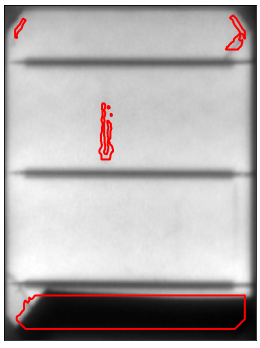}
\includegraphics[width=\x\linewidth]{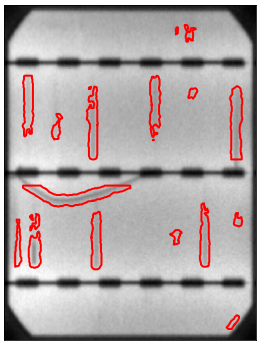}
\includegraphics[width=\x\linewidth]{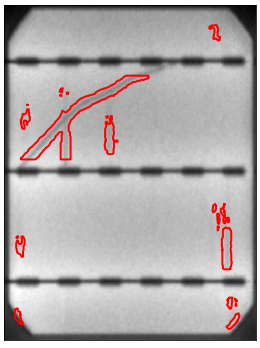}
\includegraphics[width=\x\linewidth]{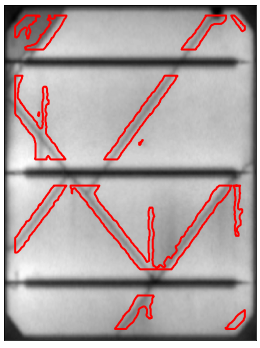}
\includegraphics[width=\x\linewidth]{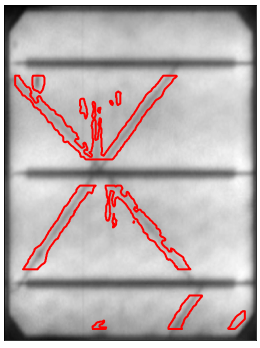}\\
\includegraphics[width=\x\linewidth]{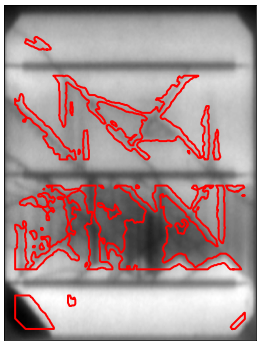}
\includegraphics[width=\x\linewidth]{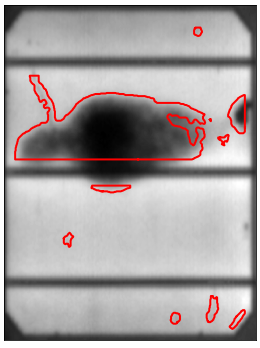}
\includegraphics[width=\x\linewidth]{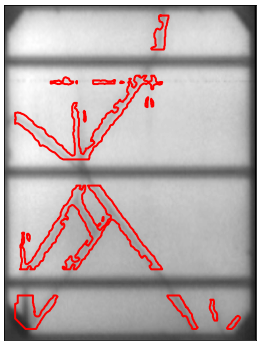}
\includegraphics[width=\x\linewidth]{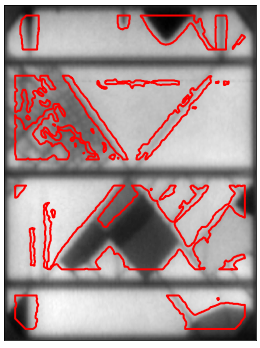}
\includegraphics[width=\x\linewidth]{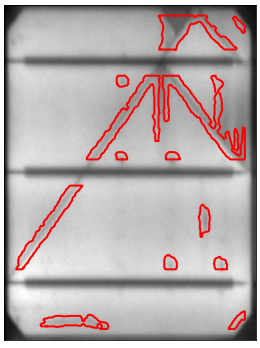}
\includegraphics[width=\x\linewidth]{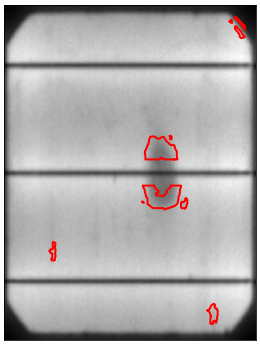}
\includegraphics[width=\x\linewidth]{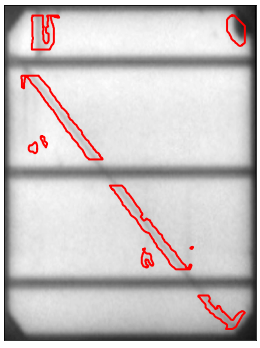}\\
\includegraphics[width=\x\linewidth]{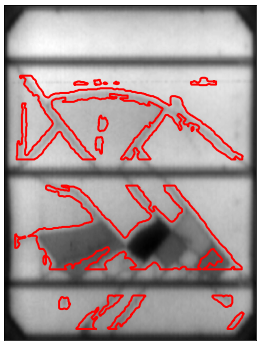}
\includegraphics[width=\x\linewidth]{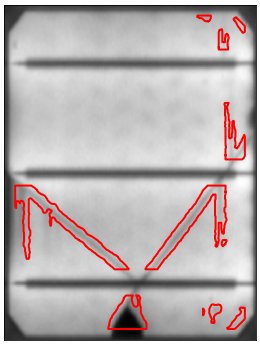}
\includegraphics[width=\x\linewidth]{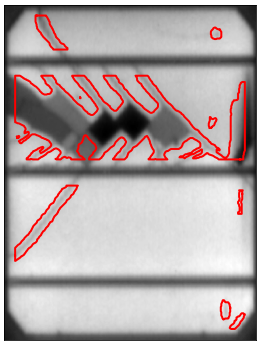}
\includegraphics[width=\x\linewidth]{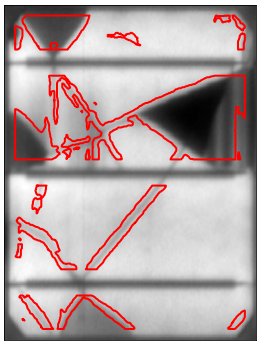}
\includegraphics[width=\x\linewidth]{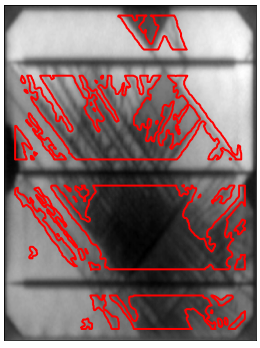}
\includegraphics[width=\x\linewidth]{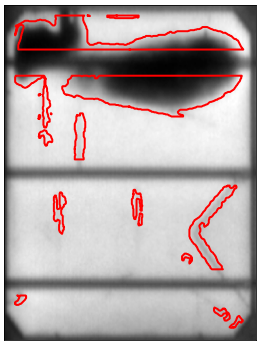}
\includegraphics[width=\x\linewidth]{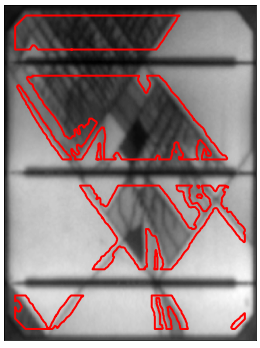}\\
\includegraphics[width=\x\linewidth]{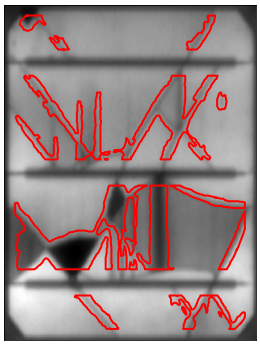}
\includegraphics[width=\x\linewidth]{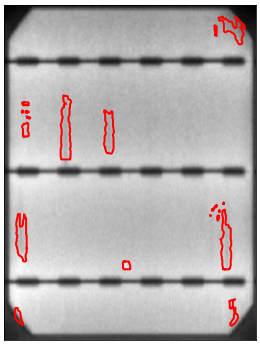}
\includegraphics[width=\x\linewidth]{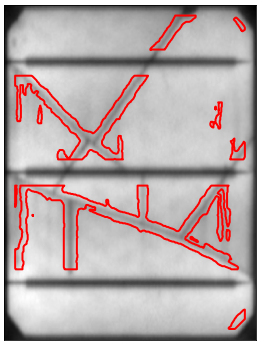}
\includegraphics[width=\x\linewidth]{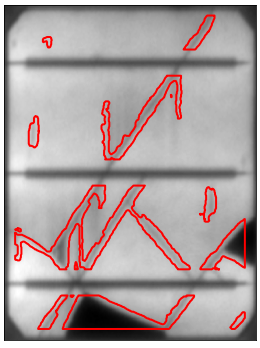}
\includegraphics[width=\x\linewidth]{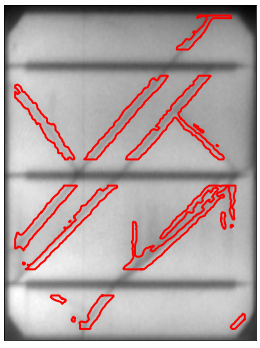}
\includegraphics[width=\x\linewidth]{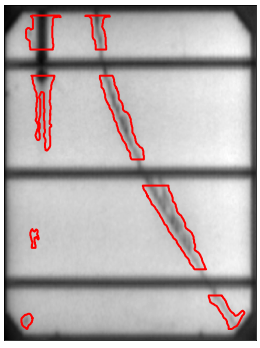}
\includegraphics[width=\x\linewidth]{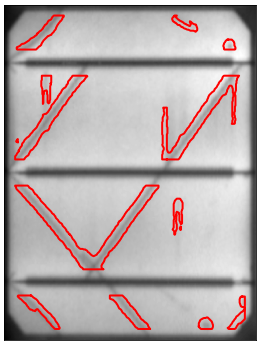}\\
\includegraphics[width=\x\linewidth]{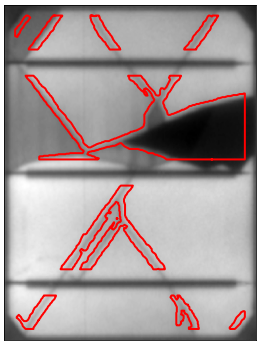}
\includegraphics[width=\x\linewidth]{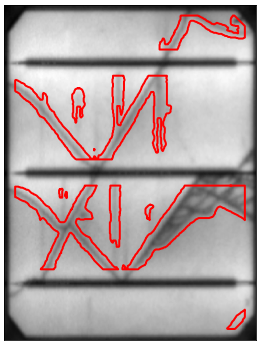}
\includegraphics[width=\x\linewidth]{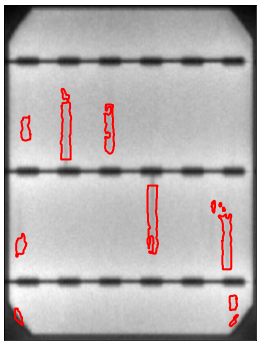}
\includegraphics[width=\x\linewidth]{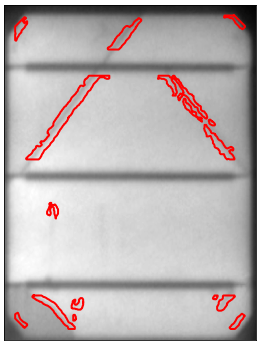}
\includegraphics[width=\x\linewidth]{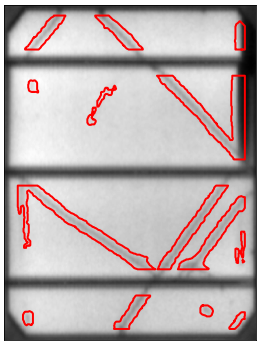}
\includegraphics[width=\x\linewidth]{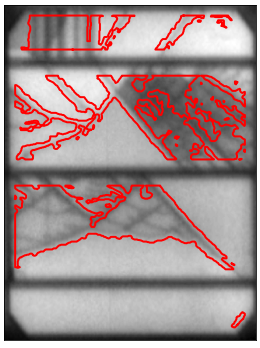}
\includegraphics[width=\x\linewidth]{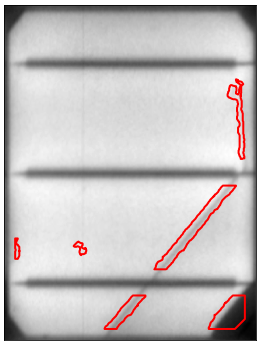}\\
\caption{ Mono-crystalline PV cells from the public ELPV dataset \cite{ Buerhop2018, Deitsch2019, Deitsch2021}. Highlighted in red are the segmentation regions detected by the proposed model. without the manual revision. }
\label{fig:appendix_a}
\end{figure*}
\vspace{-10pt}

\bibliographystyle{IEEEtran}
\bibliography{references.bib}

%


\begin{IEEEbiography}[{\includegraphics[width=1in,height=1.25in,clip,keepaspectratio]{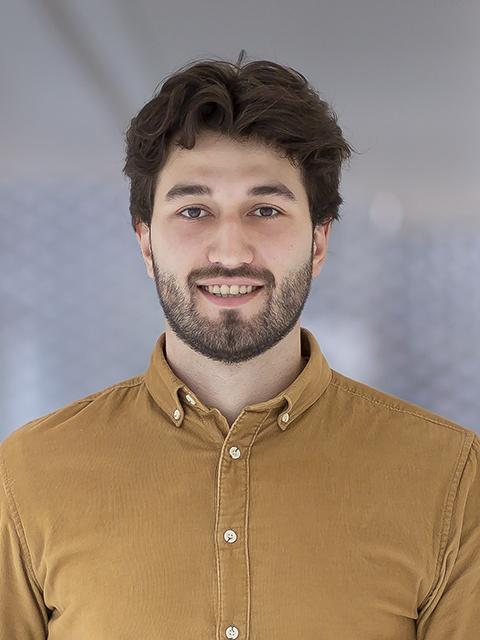}}]{Urtzi Otamendi} studied a degree in Computer Engineering at the Faculty of Computer Science of the Public University of the Basque Country (2014-2018), specializing in the branch of Software Engineering. As a final degree project, he developed a system for tracking and monitoring people using mobile devices. Subsequently, he completed the Master's Degree in Artificial Intelligence at the Polytechnic University of Madrid (2018-2019). The final master's project was carried out in collaboration with the Tekniker research center during a 5-month stay at the center. This project consisted in analyzing and implementing Deep Learning architectures for real-time detection and classification of people, vehicles, traffic signs, traffic lights, and other objects in the problem of autonomous mobility of cars. Since October 2019, he has been working as a research assistant at the Visual Interaction and Communication Technologies Center Foundation, Vicomtech, in the Data Intelligence for Energy and Industrial Processes department.\end{IEEEbiography}

\vfill

\begin{IEEEbiography}[{\includegraphics[width=1in,height=1.25in,clip,keepaspectratio]{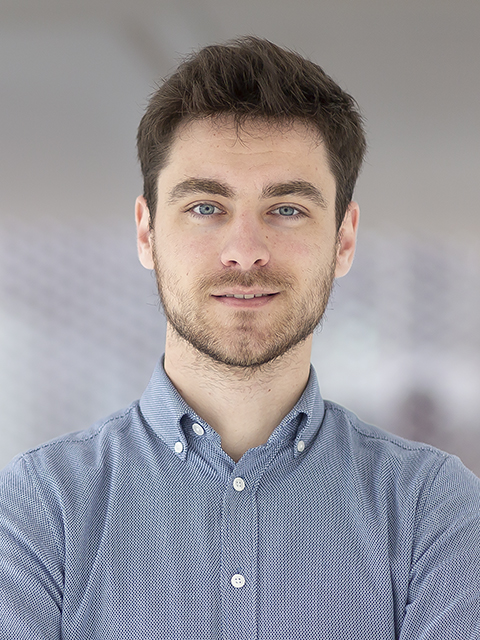}}]{Iñigo Martinez} studied a degree in Industrial Technologies Engineering at the School of Engineering of the University of Navarra (2011-2015). During his undergraduate studies, he did a research internship at the Bioinformatics department of CEIT, on the detection of cancer biomarkers using supervised learning techniques. In the summer of 2014, she carried out the final degree project in the nanomagnetism department of CIC nanoGUNE, on the design and fabrication of new magnetic materials through the deposition process. After finishing the degree, he studied the Master's degree in Industrial Engineering at the School of Engineering of the University of Navarra (2015-2017), and did the final master's project in the City Science group of the Media Lab at the Massachusetts Institute of Technology (MIT), on an active tilting system for a new generation of urban vehicles. After the stay at MIT, he worked at NEM Solutions as a data scientist, focused on early anomaly detection in time series. In March 2018 he started his PhD studies in Applied Engineering at the University of Navarra, an activity he is currently continuing. Since July 2019 he belongs to the research staff of Vicomtech as a research assistant in the Data Intelligence for Energy \& Industrial Processes department, developing projects related to data analysis, statistics and machine learning. \end{IEEEbiography}

\begin{IEEEbiography}[{\includegraphics[width=1in,height=1.25in,clip,keepaspectratio]{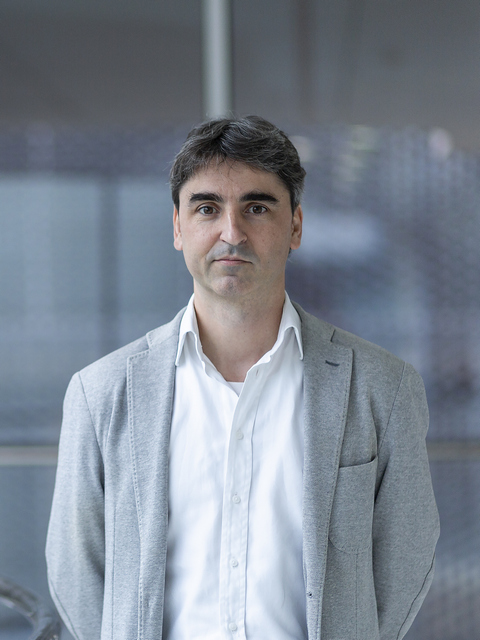}}]{Igor G. Olaizola} obtained a degree in Electronic Engineering and Industrial Automation at the University of Navarra, Spain (2001). He developed his Master thesis at the Fraunhofer Institut für Integrierte Schaltungen (IIS), Erlangen Germany (2001), where he worked for one year as a research assistant in several projects related to audio decoding according to the MPEG standard (MP3 and AAC). In 2002 he joined the research staff of Vicomtech. In 2006 he worked for a year and a half as a technology consultant in the company Vilau where he led several projects of design and deployment of Interactive Digital Television headends. In 2007 he became director of the Digital Media department at Vicomtech. In 2013 he finished his PhD at the Department of Computer Science and Artificial Intelligence of the Faculty of Computer Science of San Sebastian at the University of the Basque Country. His thesis work is based on image characterization through global descriptors and artificial intelligence techniques. Since February 2017, Igor is the head of the Department of Energy and Industrial Processes where his research group is dedicated to statistical analysis and the application of Big Data techniques. He is also currently an associate professor at the University of Navarra.\end{IEEEbiography}

\begin{IEEEbiography}[{\includegraphics[width=1in,height=1.25in,clip,keepaspectratio]{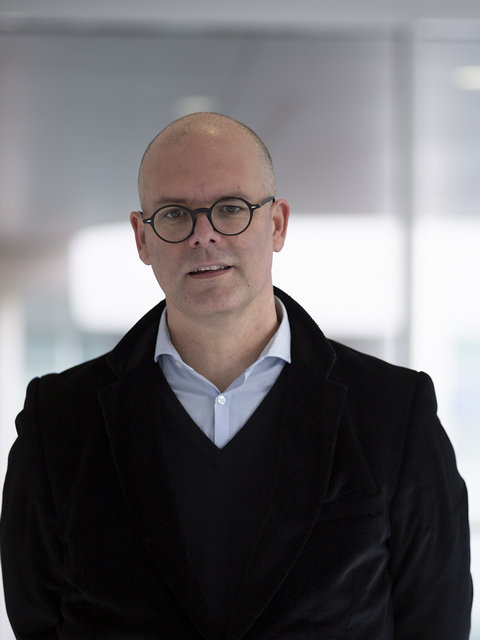}}]{Marco Quartulli} received a BS degree in Physics from the University of Bari, Italy, in 1997 and a PhD degree in EE and CS from the University of Siegen, Germany, in 2005. He worked from 1997 to 2010 in remote sensing and Earth segment engineering, image analysis and data mining at Advanced Computer Systems, Italy. From 2000 to 2003, he was with the Image Analysis Group of the Institute for Remote Sensing Technology of the German Aerospace Center (DLR). Since 2010, he has joined Vicomtech where he works in the Data Intelligence for Energy and Industrial Processes department leading data science and scalable machine learning projects. \end{IEEEbiography}

\vfill




\end{document}